\documentclass[
    hf
]{ceurart}
\usepackage[top=2.5cm, bottom=2.5cm, left=2.5cm, right=2.5cm]{geometry}

\usepackage[utf8]{inputenc}	
\usepackage{placeins} 
\usepackage{lipsum}
\usepackage{tabularx}
\usepackage{metrix}
\usepackage{lineno}
\usepackage[backend=biber]{biblatex}
\begin{document}

\copyrightyear{2024}
\copyrightclause{Copyright for this paper by its authors.
    Use permitted under Creative Commons License Attribution 4.0
    International (CC BY 4.0).}

\conference{}
\title{Metronome: tracing variation in poetic meters via local sequence alignment}


\author[1]{Ben Nagy}[%
    orcid=0000-0002-5214-7595,
    url=https://github.com/bnagy/metronome-paper,
    email=benjamin.nagy@ijp.pan.pl,
]

\author[1]{Artjoms Šeļa}[%
    orcid=0000-0002-2272-2077,
    url=https://artjomsh.github.io/web/,
    email=artjoms.sela@ijp.pan.pl,
]

\author[2]{Mirella {De Sisto}}[%
    orcid=0000-0002-0899-5976,
    email=M.DeSisto@tilburguniversity.edu,
]

\author[3]{Petr Plecháč}[%
    orcid=0000-0002-1003-4541,
    email=plechac@ucl.cas.cz,
]

\address[1]{Institute of Polish Language, Polish Academy of Sciences (IJP PAN), Kraków}

\address[2]{Tilburg University}

\address[3]{Institute of Czech Literature, Czech Academy of Sciences, Prague}

\begin{abstract}
    All poetic forms come from somewhere. Prosodic templates can be copied for generations, altered by individuals, imported from foreign traditions, or fundamentally changed under the pressures of language evolution. Yet these relationships are notoriously difficult to trace across languages and times. This paper introduces an unsupervised method for detecting structural similarities in poems using local sequence alignment. The method relies on encoding poetic texts as strings of prosodic features using a four-letter alphabet; these sequences are then aligned to derive a distance measure based on weighted symbol (mis)matches. Local alignment allows poems to be clustered according to emergent properties of their underlying prosodic patterns. We evaluate method performance on a meter recognition tasks against strong baselines and show its potential for cross-lingual and historical research using three short case studies: 1) mutations in quantitative meter in classical Latin, 2) European diffusion of the Renaissance hendecasyllable, and  3) comparative alignment of modern meters in 18--19th century Czech, German and Russian. We release an implementation of the algorithm as a Python package with an open license.
\end{abstract}


\maketitle

\section{Introduction}

All poetic forms come from somewhere, even if we tend to think they have been with us forever (as we do with familiar things). Indeed, a prosodic template can be copied in a culture for centuries or millennia; but it can also be altered by individuals, imported from foreign traditions, adapted from vernacular folksongs, or fundamentally changed by the pressures of language evolution. Yet these relationships are notoriously difficult to trace across languages and times: a lot of evidence is too fragmentary; it depends too much on written heritage, or modern ethnographic observations. National schools of metrical studies are also disparate, and rarely agree with each other, while literary scholarship over the course of the 20th century distanced itself from metrics, which are now more closely associated with linguistics and phonology. Research on the history of versification that was built on top of historical linguistics in the works of Antoine Meillet \cite{meillet_origines_1923}, Roman Jakobson \cite{jakobson_zur_1929} and, later, Mikhail Gasparov \cite{gasparov_history_1996}, was in decline, but interest is presently resurgent, following the expansion of computational and quantitative methods that allow new angles of inquiry \cite{polilova_spanish_2018, sela_semantics_2022,de_sisto_development_2023}.

This work is a continuation of the historical line of thinking about meters and metrical variation, and aims primarily to describe relationships between the \textit{form} of poetic texts. We use local sequence alignment to identify regions of structural similarities in poems; the method relies on encoding poetic texts as strings of prosodic features using a simple four-letter alphabet that can be used with any language and is independent of the organizing verse principle (quantitative, syllabic, accentual). These sequences are then aligned to derive a distance measure based on weighted symbol (mis)matches. The resulting relationships do not inherently imply any evolutionary connection between texts, but the structural similarity is illuminating: it might identify and localize emerging metrical derivatives, show continuity in forms across times and traditions, or signal formal descent, which can be useful in historical studies. While the prosodic and structural features of verse today are increasingly used for authorship attribution \cite{nagy_metre_2021, plechac_versification_2021}, this work focuses on detecting patterns that transcend individual variation and shape cultural history on a large scale: the often unseen, but powerful force of poetic meters.

Poetic meters organize speech into periodic units, causing utterances to be perceived in systemic relation to each other \cite{frog_metrical_2021}. The basis for establishing a metrical frame of reference can be different: alliteration, rhyme, recurrent prosodic patterns of vowel quality or quantity, or equivalence in duration. Meters act as rules, with varying levels of fuzziness and variation, undoubtedly constrained by cognition and language prosody \cite{rubin_memory_1995,decastro-arrazola_testing_2018,decastro-arrazola_typological_2018}. These rules, even when scholars and poets formulate them prescriptively, depend on shifting social conventions and are always tested against the affordances of a language \cite{shapir_metrum_2000}. This entails variation in the poetic realization of abstract patterns: extra-metrical stresses, additional syllables and other irregularities test the boundaries of metrical inertia and audience expectation. Metrical patterns are reaffirmed by individual usages that copy forms from the past: taken together, poems reveal governing tendencies that are not easily overwritten by individual flair. This important cultural aspect of metrical forms has prompted research into bottom-up approaches to meter recognition and verse regularity \cite{porter_space_2018, sela_measuring_2022,plechac_rhythm-based_2023} that supports clustering according to the shared structural principles that emerge from data. This descriptive approach might be considered a counterpoint to generative frameworks that abstract metrical variation under the premise of `deep structures' and language-specific `correspondence rules' \cite{fabb_meter_2008}, drawing away from the historical and cultural component of poetic meters.

We evaluate our method on the non-trivial task of meter recognition, using a labelled cross-linguistic corpus, showing the method's ability to recognize similarity in metrical organization simultaneously across different languages. Classification, however, is not our primary goal: we highlight the method's potential for historical research in three case studies that span multiple languages, versification systems and time periods (from classical Latin quantitative verse to accentual-syllabic Romantic poetry). In the hope that the methodology will find further use in the community, we release an accompanying \texttt{metronome} library for Python.%
\footnote{Available at \url{https://github.com/bnagy/metronome}, installable via \texttt{pip}.}

\section{Method}

The guiding metaphor for the development of the Metronome tools was the idea of the genome, and the four DNA molecules that are the basis for all cellular life. While consisting of only four symbols, genomes vary hugely in length and are capable of expressing minor variation, such as among human siblings, or vast differences---the same symbols define a cactus, a dolphin or a hummingbird. In the same way, the alphabet for the metronome was chosen to be as simple as possible, allowing the same kinds of analyses that are performed with DNA; comparison between very disparate samples, grouping related families into clades or clusters, and expressing or locating changes over time. We use the following symbols: for strong and weak syllables, `S' and `w', for word breaks, `.' and for end of line `|' (the specific glyphs are only mnemonics, any would work). Defining prosody in this way (strong/weak vs stressed/unstressed) allows the same alphabet to encode quantitative traditions, like classical Latin, as well as modern accentual traditions.

In bioinformatics, there are several  strategies for sequence alignment, depending on the task. For genomes (or metronomes) that are very different in length and possibly very dissimilar, the best way to compare sequences is with a `local sequence alignment'. This approach finds the shared region between two sequences that is the most similar. In contrast, a `global sequence alignment' is a better choice where the strings are closely related. We apply the well-known Smith-Waterman algorithm \cite{smith_identification_1981} for local sequence alignment, which uses a flexible alphabet and allows different bonuses and penalties for symbol (mis)matches. Smith-Waterman is agnostic regarding the sequence alphabet---it is commonly applied both to nucleic acid sequences (four symbols) as well as proteins (20 symbols), and produces a match score that depends on a bonus/penalty matrix. In our work, the best local match score is normalized using the self-score of the shorter sequence, producing a comparable distance (or similarity) in [0,1].

The general intuition behind the bonuses and penalties (which determined the initial values) was based on empirical understanding of poetic meter---for example matching line lengths are rewarded more than matching syllable strengths, and mismatches between word breaks and syllables are not penalized at all (although matching word breaks are rewarded, which allows the algorithm to group schemes that have customary caesura positions, discussed more in the Appendix). To optimize the bonus/penalty matrix, we then performed a series of iterative classification tasks on the cross-language dataset, and visually inspected the clustering results of several corpus subsets with known metrical schemes.

The metronome package leverages methods from the BioPython package \cite{biopython} which provides highly optimized routines for the sequence alignments. Our code adds an assortment of sampling strategies, with the core of the package being a parallel method that produces an $n \times n$ distance matrix from n input samples (which have been encoded as metronomes). These distance matrices can be used directly as inputs to create hierarchical clustering / dendrogram methods (such as \texttt{hclust} in R) or to spatial clustering methods like UMAP (using a `pre-calculated' metric).

\section{Evaluation}

We evaluate the performance of the algorithm in a non-trivial cross-linguistic task by looking at clustering performance on samples of individual poems in twelve meters, and coming from four languages (Czech, German, Russian, and Classical Latin). We compare this performance to three baselines: simple edit distance, local alignment with an unweighted substitution matrix, and a machine-learning approach based on the frequencies of prosodic $n$-grams.

\subsection{Data and sampling}

Our modern language subset (67,940 works) covers the six most common meters: iambic and trochaic pentameter, tetrameter, and tetrameter/trimeter (alternating lines of tetrameter and trimeter, sometimes called `common meter' in English). The Czech subset (44416 works), German subset (15172 works), and Russian subset (8352 works) come from the PoeTree dataset  \cite{plechac_poetree_2023}. Metrical annotation in Czech was performed using machine learning methods, and the output was manually verified  \cite{plechac_czech_2016}; in German it was produced using a rule based algorithm \cite{bobenhausen_metrique_2015}, and a similar approach was implemented in Russian, and used the accentual classes of words to infer rhythm.

The Classical Latin subset (2588 works) covers the six most common meters from that corpus: elegiac couplets, hendecasyllables and stichic hexameter (making up the bulk of samples and authors), as well as a few senarii, scazons, and alcaics. The largest language / meter set is Czech iambic pentameter (over 17,000 works) and the smallest is Latin alcaics (38 works, all from Horace).

Since the samples are of such disparate size, we downsampled to just 10 examples of each language / meter combination, yielding 240 random samples per evaluation run. We then classified samples by meter (twelve meter labels) which implies a random baseline accuracy of about 8\%.

\begin{figure*}
    \includegraphics[width=\linewidth]{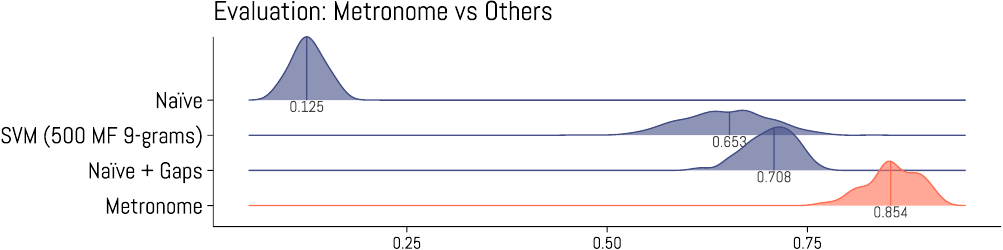}
    \caption{Classification performance of Metronome vs SVM and two baseline alignment algorithms. Smoothed distribution of accuracy results over 50 random subsamples, with median scores.}
    \label{fig:eval}
\end{figure*}

\subsection{Classification Methods}

For the alignment-based methods, which operate in a similar way to Metronome, we created an $n \times n$ distance matrix, to which we applied a $k$-Nearest Neighbours (kNN) classifier ($k$=7), with a simple accuracy metric. These alignment-based methods are as follows. For the Naïve sequence aligner, we take the best local alignment, but without permitting indels (insertions or deletions inside the aligned areas). This is similar to a local Hamming distance, and is a very unforgiving alignment (since word breaks are included), used only as a baseline. Next we added indels to the Naïve aligner, making it more like a local Levenshtein distance but with uniform match/mismatch bonuses and penalties---this is similar to the way local sequence aligners would work on DNA sequences. Finally, we used the full Metronome configuration with a variable penalty matrix. The fourth method, and the only one that was not alignment-based, was a standard machine-learning approach, in which we applied an SVM classifier to the frequencies of the 500 most-frequent symbol 9-grams, using an 80/20 train/test split. 9-grams were chosen because the symbol alphabet is very small, and so it is advantageous to the classifier to be able to match almost an entire line. Based on testing, the best results were with 500 most-frequent 9-grams, reduced by SVD to 50 dimensions and then $z$-scaled. It must be noted here that the SVM classifier is hampered greatly by the presence of word-breaks. For most meters, word breaks are not part of the metrical constraints, they are only relevant to some questions of style and in a few traditions that have mandatory or customary caesurae in certain places (it is very important in classical Latin and also in modern alexandrines, for example). In other words, in terms of predicting the name of the metrical pattern, the word breaks are noise. If the aim were simply to create an optimal classifier, machine-learning approaches perform at least as well as Metronome (and are much faster), at the expense of nuance and interpretability, by simply discarding the word breaks.

\subsection{Results}

The classification tests were run 50 times each, and the full distributions and medians are reported in Figure \ref{fig:eval}. As discussed above, the unexpectedly poor performance of the SVM classifier (usually a solid benchmark) is mainly due to the presence of word breaks, but it is also clear that the custom penalty matrix used in Metronome provides a solid increase in performance across languages and traditions, compared to the standard string aligner. Classification accuracy, however, is merely a proxy for the real utility of Metronome as a tool to understand metrical \textit{relationships} between poems. In the following section we provide a few small examples of this potential.

\begin{figure*}
    \includegraphics[width=\linewidth]{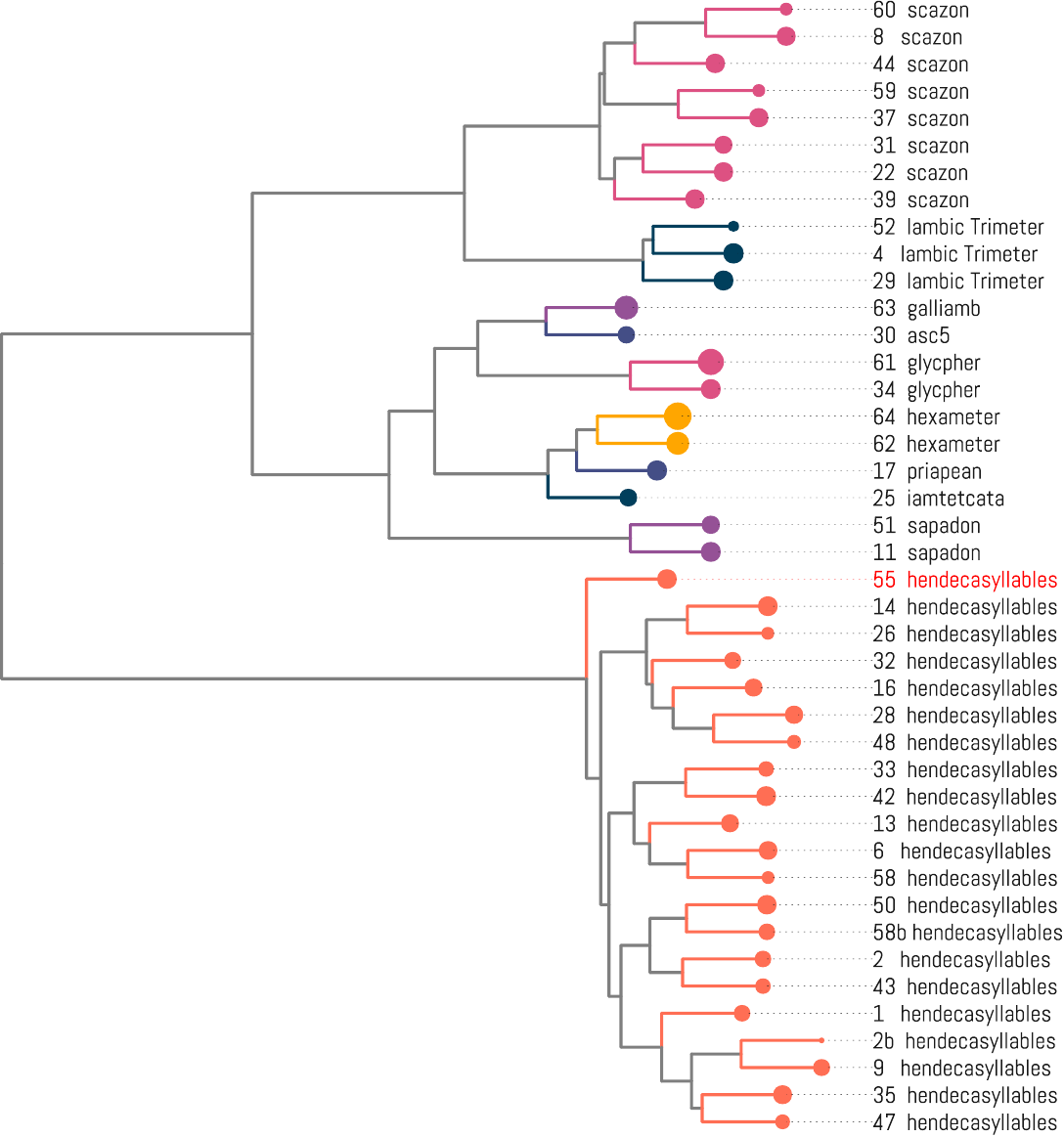}
    \caption{Cladogram of a selection of poems by Catullus. \emph{Carmen} 55, while still composed in hendecasyllables, is visibly different to the rest of that clade.}
    \label{fig:cat_phylo}
\end{figure*}

\begin{figure*}
    \centering
    \includegraphics[width=0.8\linewidth]{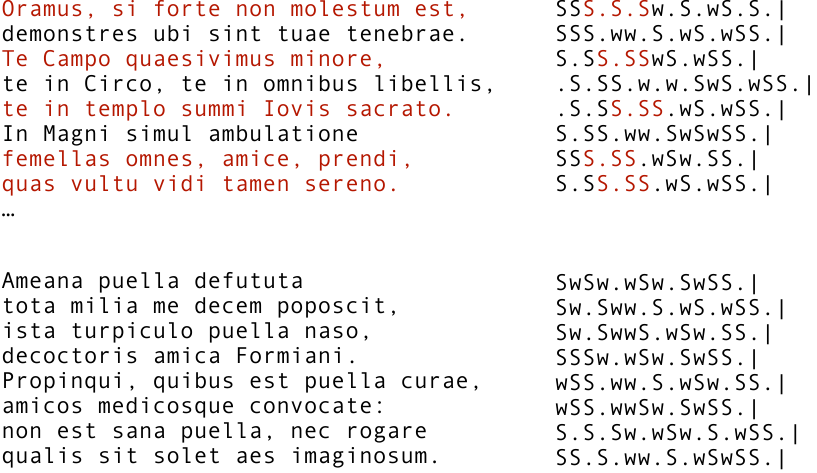}
    \caption{A visual comparison of the metronome strings (formatted to add line breaks) for the beginning of \emph{Carmina} 55 (variant with collapsed choriamb) and 41 (standard hendecasyllable).}
    \label{fig:55_41}
\end{figure*}

\section{Showcases}

\subsection*{Showcase A: Mutations in Classical Latin meter}

The first showcase simply highlights the sensitivity of the metronome to minor metrical variations. During a manual inspection of the dendrogram of the poems of Catullus, it was observed that one hendecasyllable poem, Catullus \textit{Carmen} 55, was quite different to all of the others of that meter. This can be seen in Figure \ref{fig:cat_phylo}---notice how poem 55 branches off from the rest of the hendecasyllable group much `earlier' (indicating more difference), and yet is still clearly part of the overall clade. Thinking that this might be an algorithmic error, we inspected the text, and discovered that 55 is a metrical oddity. In 55, Catullus often collapses the central double breve of the choriamb (\metricsymbols{_ uu _}) to form a mollossus (\metricsymbols{_ _ _}), yielding a line of ten syllables instead of the standard 11 (hendecasyllable is, literally, `11 syllables'). This variation can be seen in Figure \ref{fig:55_41}, which shows the start of 55, alongside its metronome transformation, contrasted with 41, a normal hendecasyllable with a standard choriamb in every line. This license occurs only twice more in all of Catullus (in 58b), making 55 a very odd animal indeed.%
\footnote{An approachable description of the meters of Catullus, including a discussion of this issue, can be found in Peter Green's bilingual edition  \cite{catullus_poems_2005} p. 32 ff.}
It is precisely this kind of `loose' family grouping, based on the meters as they are written, not as they are defined by textbooks, that highlights the exciting potential of the metronome as a tool to provide insights from a `distant reading' of poetic meter.

The raw data for this showcase were provided by David Chamberlain, who maintains an extensive digital selection of fully scanned Greek and Latin verse under an open license \cite{hypotactic}. Minor post-processing was all that was required to convert it to the Metronome string format.

\subsection*{Showcase B: European diffusion of a `Renaissance meter'}

\begin{figure*}
    \includegraphics[width=\linewidth]{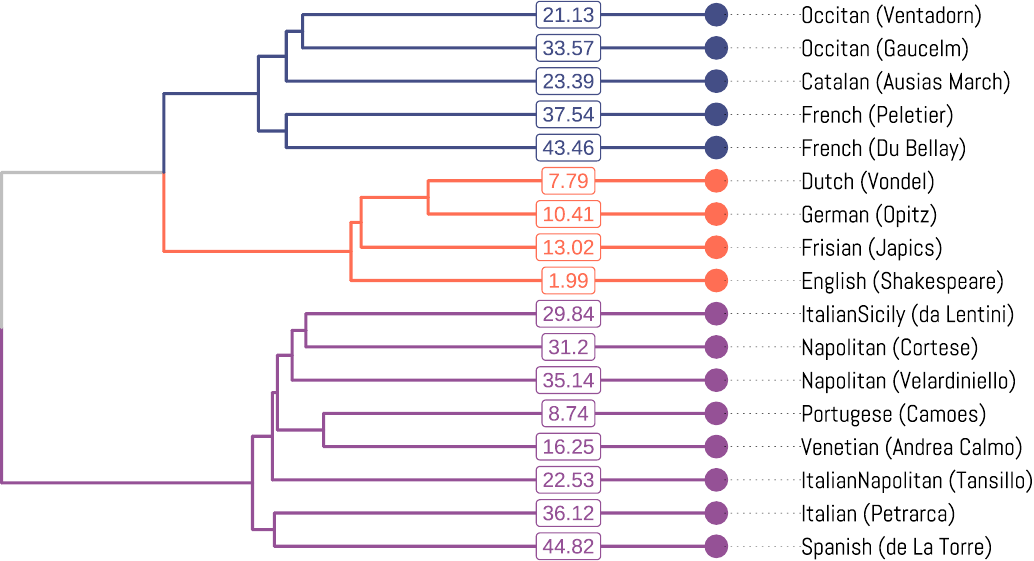}
    \caption{A metronome-based cladogram of various samples of Renaissance meter. The inset number is the entropy-based variability from the regular metrical form (see \cite{sela_measuring_2022}). Shakespeare is the most regular, de La Torre the least.}
    \label{fig:ren_dendro}
\end{figure*}

The second showcase focuses on a typological comparison of different implementations of Renaissance meter. During the Renaissance, a new metrical form  spread across Europe by replacing previous forms, or by causing readjustments to the pre-existing templates. Originally it was a regular 10-syllable line that became 11-syllable in Italian because of its fixed penultimate stress---this hendecasyllable became widely known through Petrarch's poetry. Later, the same meter developed into the staple form of modern English: the iambic pentameter.

In each tradition that adopted the innovation, the meter underwent some changes and adaptations to the recipient language and versification style which led to different implementations of the same poetic form \cite{de_sisto_interaction_2020}. The new meter probably originated in the Occitan tradition \cite{beltrami_cesura_1986, di_girolamo_i_1999, billy_linvention_2000} and first reached Italy from there. The Europe-wide renown of Petrarch and his \textit{endecasillabo} quickly spread the new meter to other Romance traditions, like Spanish, and vernaculars of the Italian peninsula, like Neapolitan, Venetian and Sicilian. French poetry was already using decasyllabic verse (\textit{amour courtois}) \cite{hudson_short_1919}, but this form was adjusted to the Italian trend of composing sonnets in \textit{endecasillabo} \cite{hudson_short_1919,key_short_2006}, before being replaced by the alexandrine meter. Occitan verse also influenced Catalan and Portuguese poetry, which already had a pre-existing decasyllabic form.

Later in the Renaissance, the new metrical form spread independently to English and to Dutch poetry. Subsequently, the Dutch Renaissance meter influenced German and Frisian poets. All West Germanic adaptations developed strict iambic meter from the contact with isosyllabic forms. This was the beginning of modernity in European poetry: the rise of accentual-syllabic meters everywhere east of France.

The metrical alignment (see Figure \ref{fig:ren_dendro}) of small samples from many Renaissance traditions clearly distinguishes between the three previously described `directions' of adoption of the form: 1) Italian peninsula and Spanish; 2) Occitan, Catalan and French; and, finally, 3) Germanic traditions that pivoted to regular iambic meter (the values on the graph show how `free' the samples are with regard to the placement of stresses---note how the Germanic samples are close to 0). While the finer historical relationships between traditions may remain opaque to this method (which depends on the individual samples, and technical issues with word boundaries like synalepha),%
\footnote{
    Poetry in many Romance traditions (Spanish, Italian \textellipsis) makes extensive use of synalepha---merging two syllables into one---which complicates syllable counting and word boundary placement. We manually annotated all instances of synalepha, and treated cross-word syllable elision as one `phonetic word', placing a word boundary after the contracted sequence. For metronome analysis, all synalephas were annotated as a single syllable: ``strong'' if there was a strong syllable present on either side of the elision, and ``weak'' when both syllables were unstressed.
}
overall it captures the key structural and linguistic similarities emerging from the implementation of the same form.

The data employed in this showcase were originally compiled by Mirella De Sisto \cite{de_sisto_interaction_2020}; the samples were selected from the works of poets who can be considered representative of the Renaissance tradition of their respective languages. Manual annotation was performed, in collaboration with experts in the phonology and metrics of the respective languages.

\subsection*{Showcase C: Modern expansion: accentual-syllabic verse}

During the 18--19th centuries, foot-based accentual-syllabic verse, which emerged through interaction with isosyllabic
Renaissance meter, rapidly spread from Germany to the North and East \cite[207-209]{gasparov_history_1996}
\cite{kazartsev_iambic_2022} and was critical for establishing newly emerging national literatures. Through highly
influential (and highly edited) folksong collections  \cite{abrahams_phantoms_1993,leerssen_oral_2012}, like the
\textit{Volkslieder} (1786), regular trochaic meters became closely associated with oral traditions throughout Europe,
while iambic verse, from German courtly odes to Shakespeare's dramatic genius, provided a basis for high-status
secular literature. This formed a cornerstone for the  enduring cultural---and, as a result, semantic---opposition of
iambic and trochaic meters \cite{sela_semantics_2022}.

As with any technology, conquest and power played a major role in metrical adoption, and it is not a coincidence that `fashionable' accentual-syllabic verse spread quickly through the territories, communities, and languages of the Austro-Hungarian and Russian empires, also following the unification of literary languages and school systems \cite[238-239]{gasparov_history_1996}. New verse often disrupted already existing versification, or weakened local alternatives (for example, purely syllabic verse in Polish, Czech and Ukrainian).

\begin{figure*}
    \includegraphics[width=\linewidth]{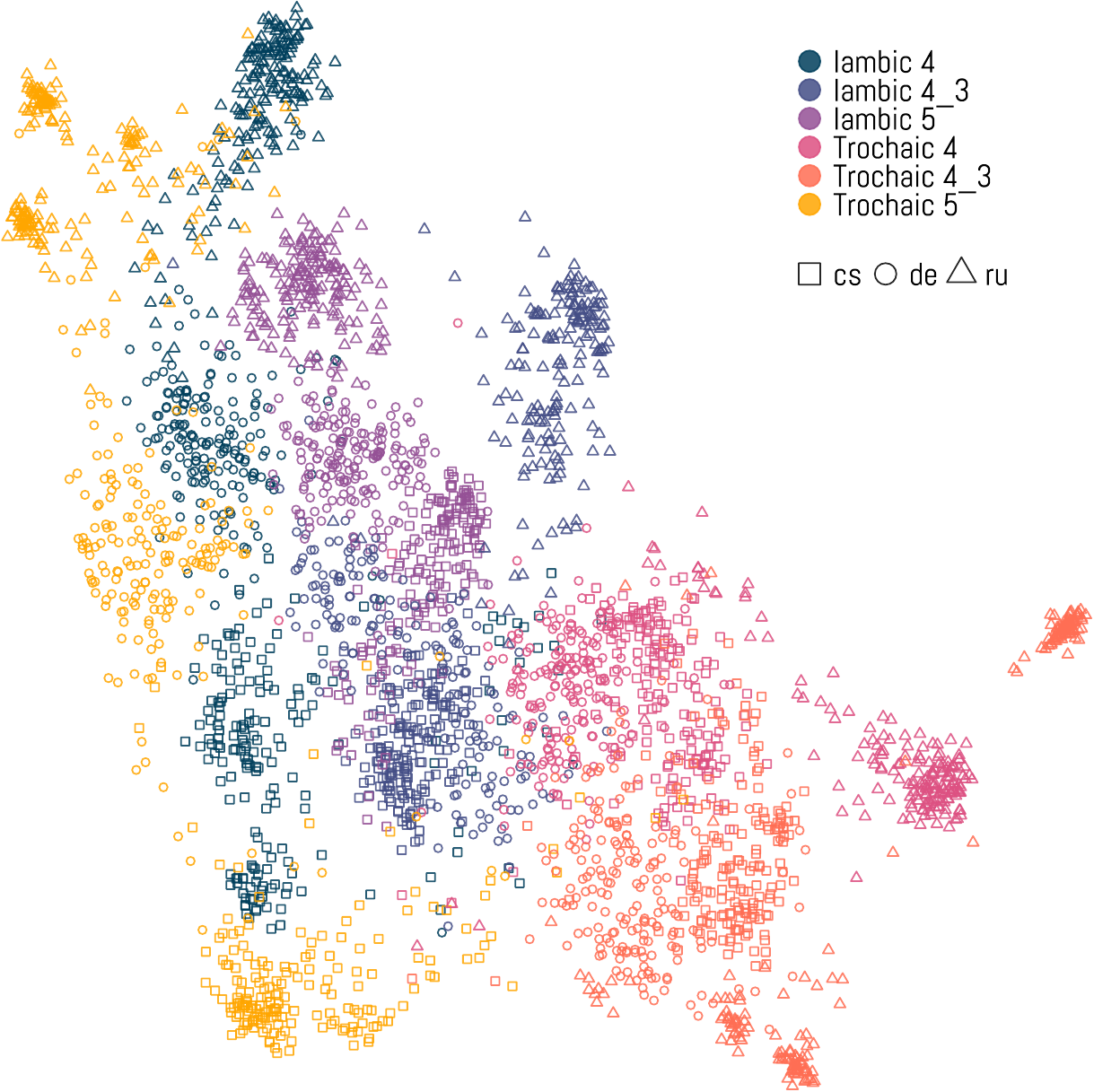}
    \caption{UMAP cluster of 3222 poems in Czech, German, and Russian from the PoeTree corpus, in the six most common European meters. Metronome distance is used as the clustering metric.}
    \label{fig:crosslang}
\end{figure*}

The last showcase (Fig. \ref{fig:crosslang}) demonstrates the possibility of cross-linguistic alignment of distinct modern poetic forms in Czech, German, and Russian. We are able to recognize the structural unity of the same meters (colors), detect the broader divide between trochee and iamb (inner and outer regions on the plot), yet preserve language-specific variation (point shapes)---the latter occurs partly because of differences in rhythm, and partly because word boundaries also encode differences in word lengths. German verse was a source for both Slavic literatures at different times, and the evidence here of structural similarity is a step towards better understanding the individual metrical lineages, and tracing formal connections through global literary history.

The data presented in Fig. \ref{fig:crosslang} is a subset of the same data we used in our evaluation (cross-linguistic meter recognition) and comes from annotated corpora of Czech, German, and Russian poetry (all parts of the PoeTree).

\section{Discussion}

We move now to consider some limitations, and opportunities for further work. Our approach was informed by meters that depend on isosyllabicity and recurrent prosodic patterns, so we expect it to struggle with forms that emerged around other principles. For example, alliteration patterns that define verse in Old English and Old Norse (\textit{dróttkvætt}) cannot be traced and would require a radically different approach to encoding the `alphabet' for alignment. The same can be said of poetic forms in the same meter that are distinguished based on the shape of stanza and/or a pattern of rhymes. For example, both \textit{ottava rima} and English sonnets can be written in iambic pentameter, yet represent distinct traditions. In this case, the expansion of the alphabet would be straightforward: adding a symbol for a stanza boundary would allow us to trace basic stanza composition (but make the analysis less broadly applicable). Additionally, the general methodology of Metronome (alignment of abstracted schemes) might be applied to rhyme patterns instead of rhythm sequences: one can envision, for example, a reconstruction of the history and typology of European sonnets based on classic rhyme-to-letter encoding (abab|abab|cdc|cdc).

Our current approach is also under-investigated in relation to caesurae---fixed-position word boundaries that can play a significant role in defining meters, like the French alexandrine (a meter of 12 syllables, where a caesura consistently bisects the verse, 6+6). To address this, we performed metronome analysis on simulated variations of the alexandrine (see the Appendix), to see if caesura-based forms can be distinguished from plain syllabics in our current approach. Preliminary results are very promising: even under conservative restrictions, the metronome tends to distinguish pseudo-poems that are governed by caesurae vs. pseudo-poems of the same line-length that are not.

Finally, the method depends on the availability of raw scansions of large collections of poetry. This can be tricky, since scansion itself depends on existing metrical theory, shared within a community, a consensus that is rarely maintained across national traditions of scholarship. On one hand, theory-agnostic scansion is hardly ever possible, but on the other hand, the presence of theory in automated scansion systems can create self-fulfilling cycles, where empirical patterns are forced to fit a few anticipated schemes (for review see \cite{de_sisto_understanding_2024}), which regularizes and conceals actual variation.

In the authors' other work we often advocate for the utility  of bottom-up approaches to meter recognition, since any
pattern that is repeated enough times, any strong organizational principle, will always make itself visible; the very
nature of systematic meter will cause it to leave its mark on any observation or measurement, be it taken on a text's
prosody, morphology, or syntax \cite{gasparov_linguistics_2008}. Meters are, essentially, quite simple technologies of
rule iteration: the more you repeat one, the more obvious it becomes, even for such crude measuring instruments. This
makes even a `sub-optimal' automated scansion, made without expert knowledge of the underlying theory and tradition,
usable and useful.

In conclusion, this study shows how the alignment of metrical strings can be enlightening in a variety of contexts---from tracing slight deviations in well-established meters, to recognizing large groups of similar forms across languages and times. Our proposed method works well with both long and short sequences, and is applicable in a wide range of poetic traditions (our showcases feature three major ones: quantitative, syllabic and accentual-syllabic). The method's sensitivity to large and small structural differences in verse organization highlights its potential in both historical and cross-linguistic comparative research.

\section{Availability of Data and Code}\label{sec:data}

The preprint repository may be found at \url{https://github.com/bnagy/metronome-paper}. All code
and data is available under CC-BY, except where restricted by upstream licenses.
The code repository includes full reproduction data and code for the evaluation,
as well as various supplemental figures and explanations.

\FloatBarrier

\begin{acknowledgments}
    Ben Nagy is supported by Polish Academy of Sciences Grant 2020/39/O/HS2/02931. Artjoms Šeļa was supported by the
    project ``Large-Scale Text Analysis and Methodological Foundations of Computational Stylistics'' (SONATA-BIS
    2017/26/E/HS2/01019). Mirella De Sisto was supported by the research project CIGE/2022/114 funded by the Valencian
    Government (GE 2023, Conselleria d'Innovació, Universitats, Ciència i Societat Digital). Petr Plecháč was supported
    by the project ``European Poetry: Distant Reading (2023--2025)'' (Czech Science Foundation GA23-07727S). Data
    visualisations are produced with \texttt{ggplot} \cite{ggplot} and cladograms use \texttt{ggtree} \cite{ggtree}.
\end{acknowledgments}

\printbibliography
\FloatBarrier

\appendix
\section*{Appendix. Simulated alexandrines}\label{sec:appendix}

\begin{figure*}
    \includegraphics[width=\linewidth]{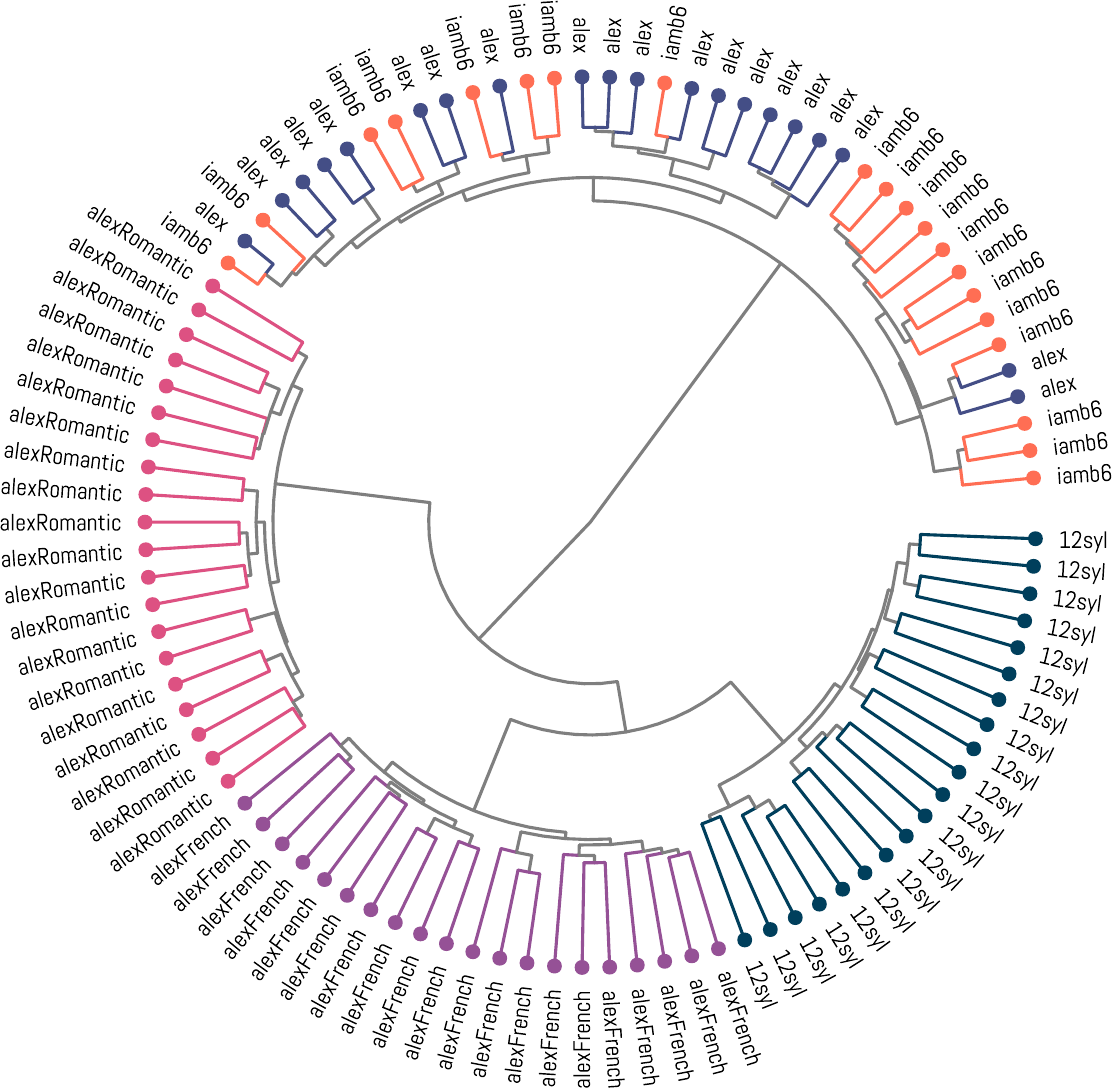}
    \caption{Clustering of simulated pseudo-poems that represent five conditions of the alexandrine form: three syllabic and four accentual-syllabic}
    \label{fig:sim_alex}
\end{figure*}

Our analysis of modern meters in this paper did not require the detection of caesurae---systemic word boundaries in the line (and often a syntactic pause) that help to define the meter in some traditions, like syllabic French verse and classical Latin. The standard modern example is the French 12-syllable alexandrine with a central caesura following a strong syllable. It is important to know whether the metronome can distinguish forms based on the presence, or placement, of caesurae. To test this, we generated pseudo-scansion sequences under four different conditions that imitate, albeit naïvely, verse prosody.

Each generated poem has a size drawn from a Poisson distribution with $\lambda=14$, so that the length of all poems in the synthetic tests will average 14 lines, the length of a sonnet.

\begin{enumerate}
    \item \textbf{Iambic hexameters with alexandrine caesura (alex)}: each line is strictly a sequence of ``wSwSwS.wSwSwS'', with an obligatory word boundary after the 6th syllable. All other word boundaries in the line are determined randomly from a distribution of probable word-lengths;
    \item \textbf{Plain iambic hexameter (Iamb-6)}: same as above, but no hard-coded caesura, word boundaries are assigned randomly;
    \item \textbf{Classic French alexandrine (alexFrench)}: each line has a general structure of ``xxxxxS.xxxxxS'', where x is any (w or S) symbol, and word boundaries are determined freely in each hemistich.
    \item \textbf{`Romantic' alexandrine (alexRomantic)}: has a general structure of xxxS.xxxS.xxxS; this form is associated with Victor Hugo and Romantic experiments with the classic meter;
    \item \textbf{Plain 12-syllable meter (12syl)}: a hypothetical meter that does not have any restrictions, except line length.
\end{enumerate}

We generate lines under simple assumptions and do not follow the structure of any language in particular, nor do we try to plausibly reproduce word-length distribution and stress placement within words (so e.g. unlikely words with prosodic structures like ``ww'' or ``SSS'' can occur). To distribute word boundaries within a line,  we draw 1-,2-,3-, and 4-syllable words with the corresponding probability ratios 1:3:1:0.25. This ratio leads to a  probability of ~45\% that there will be a word boundary after the 6th syllable in a non-alexandrine line, simply by chance. This is reasonably high, so the metronome needs to prove its sensitivity in tracing regular differences.

While this setup is unrealistic, it is transparent, and allows us to control metrical principles independent of linguistic regularities. Figure \ref{fig:sim_alex} shows the resulting clustering of the `poems': it is clear that the forms that employ caesurae can be distinguished from those that do not. The alignment works better for `syllabic' alexandrines because, unlike iambics, they have an additional distinctive feature---two fixed stressed positions versus an unregulated 12-syllable line.

Further tests confirm the outcome: for 100 runs with 20 poems generated for each form, we measure the cluster efficiency via Adjusted Rand Index (ARI). The median ARI for iambic meters is 0.55, indicating a somewhat noisy recognition of caesura-based forms. The median ARI for syllabics is 0.95---an almost perfect clustering of the three varieties. Thus, we expect that the main structuring principle of caesura-based forms should be discoverable, even in very noisy settings.

\onecolumn

\end{document}